\PassOptionsToPackage{table}{xcolor}
\documentclass{article}

\usepackage[preprint]{corl_2023} 

\usepackage{tabu}
\usepackage{graphicx}
\usepackage{wrapfig}
\usepackage{subfig}
\usepackage{tabularx}
\usepackage{listings}
\usepackage{amssymb}

\definecolor{codegreen}{rgb}{0,0.6,0}
\definecolor{codegray}{rgb}{0.5,0.5,0.5}
\definecolor{codepurple}{rgb}{0.58,0,0.82}
\definecolor{backcolour}{rgb}{0.95,0.95,0.92}

\title{Tell Me Where to Go: A Composable Framework for Context-Aware Embodied Robot Navigation}

\author{
  Harel Biggie\\
  Department of Computer Science\\
  University of Colorado Boulder \\
  \texttt{harel.biggie@colorado.edu} \\
  \And
  Ajay Narasimha Mopidevi \\
  Department of Computer Science \\
  University of Colorado Boulder \\
  \texttt{ajay.mopidevi@colorado.edu} \\
  \And
  Dusty Woods\\
  Department of Computer Science \\
  University of Colorado Boulder \\
  \texttt{destin.woods@colorado.edu} \\
  \And
  Christoffer Heckman \\
  Department of Computer Science \\
  University of Colorado Boulder \\
  \texttt{christoffer.heckman@colorado.edu} \\
}

\begin{document}
\maketitle

\vspace{-15pt}

\begin{abstract}
Humans have the remarkable ability to navigate through unfamiliar environments by solely relying on our prior knowledge and descriptions of the environment. For robots to perform the same type of navigation, they need to be able to associate natural language descriptions with their associated physical environment with a limited amount of prior knowledge. Recently, Large Language Models (LLMs) have been able to reason over billions of parameters and utilize them in multi-modal chat-based natural language responses. However, LLMs lack real-world awareness and their outputs are not always predictable. In this work, we develop NavCon, a low-bandwidth framework that solves this lack of real-world generalization by creating an intermediate layer between an LLM and a robot navigation framework in the form of Python code. Our intermediate shoehorns the vast prior knowledge inherent in an LLM model into a series of input and output API instructions that a mobile robot can understand. We evaluate our method across four different environments and command classes on a mobile robot and highlight our NavCon's ability to interpret contextual commands.
\end{abstract}

\keywords{Natural language, navigation, contextual navigation} 


\section{Introduction}


Humans have the remarkable ability to navigate through unfamiliar environments, e.g., in a town or through a building, by relying solely on our priors and descriptions of the environment \cite{mcnamara1989subjective}. Motivated by the difficulties in directing robots in collaborative teams with humans such as those used in search and rescue \cite{DARPA2022, agha2021nebula, biggie2023flexible, tranzatto2022cerberus} operations, we aim to develop a framework that allows humans to provide high-bandwidth instructions to robots in the form of natural language.

To achieve this, it is generally required that a robot would associate natural language utterances to the physical world using sensing modalities onboard the robot, a process known as grounding. However, unlike humans, robots are not currently capable of integrating prior experiences into a vast wealth of priors to aid this association.

\begin{figure}[htb]
    \centering
    \includegraphics[width=\textwidth]{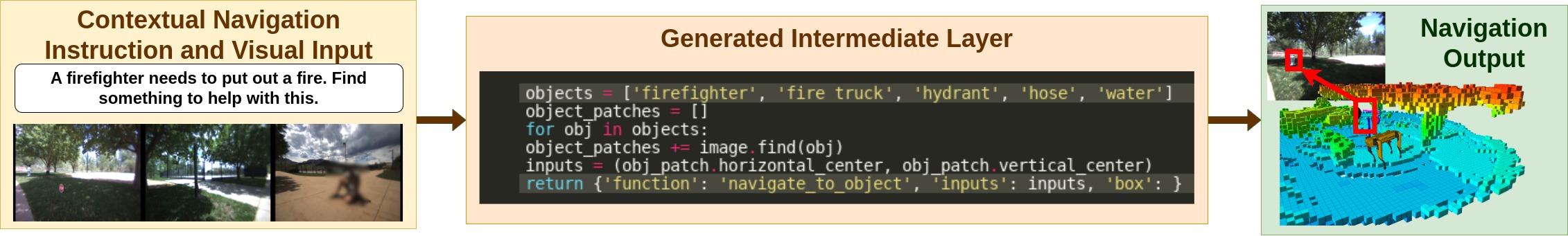}
    \caption{Contextual navigation example for a firefighting situation. We see our compositional framework generates code for the robot to both find and then plan to a fire extinguisher.}
    \label{fig:firefighter}
\end{figure}

Recently, Large Language Models (LLMs) \cite{zhao2023survey} have been able to reason over billions of parameters and utilize the results in domains such as dialogue-based responses \cite{glaese2022improving, tsimpoukelli2021multimodal, liu2023pre}, code generation \cite{chen2021evaluating, li2022competition, xu2022systematic}, and multimodal visual question and answering (VQA) tasks \cite{alayrac2022flamingo}. This type of unsupervised learning shows incredible potential for the generalization of prior knowledge but these models lack real-world experience. A lack of physical experience is perhaps one reason why transitioning techniques developed through research in VQA and view-based navigation into embodied agents has not kept pace with the research in these fields.

Additionally, with the current generation of LLMs, it's inherently challenging to interpret the rationale behind the outputs that are generated by the model. In this work, we address both the grounding and transparency issues using a composable framework designed for robot navigation. Our pipeline extends the reasoning framework for visual inference presented in \cite{suris2023vipergpt} to an embodied robotic agent. With our composable framework, we leverage LLM priors (GPT-3.5), state-of-the-art object detectors \cite{li2022grounded}, and classical robotic planning algorithms \cite{lavalle1998rapidly, dang2020graph, ahmad2022efficient} to perform zero-shot natural language based navigation in four unique environments.

Specifically, our contributions are as follows:
\begin{itemize}
    \item We extend the concept of modular neural networks and define a new framework for composable robot navigation. Our framework requires a minimal uplink for the robot since all of our planning, mapping, and localization is performed onboard.
    \item We evaluate different 2D input representations to determine an effective way to extract spatial and conceptual knowledge from LLMs.
    \item We perform extensive real-world experiments in a variety of environments and show that our framework is able to navigate to landmarks based on natural language. Furthermore, the framework is able to deduce appropriate navigational goals from the context of a sentence.
\end{itemize}



\section{Related Works}
\label{sec:related_works}

\textbf{Grounding Language:} Associating natural language with an embodied agent requires grounding utterances to the physical world in which the robot operates. Various approaches have been used to associate language with the physical domain ranging from probabilistic graph-based structures \cite{tellex2011understanding,howard2014natural,kollar2010toward} to end-to-end learning-based methods \cite{driess2023palm}. Graph-based approaches have shown promising results but their generalizability is limited to a fixed training corpus. On the other hand, LLMs have proven to be adept at reasoning over large amounts of unsupervised training data \cite{floridi2020gpt, touvron2023llama} using transformer-based backends \cite{vaswani2017attention}. These models also known as foundational models due to their comprehensive knowledge have recently been leveraged for task and motion planning using reinforcement learning on a set of skills \cite{brohan2023can}, and in an end-to-end fashion \cite{driess2023palm} with object scene representation transformers \cite{sajjadi2022object}. While these approaches have made remarkable strides they still suffer from the explainability problem and further they don't share any of the path planning guarantees that traditional planning methods share \cite{noreen2016optimal}.

\textbf{Modular Neural Network Frameworks and Code Synthesis: } Modular Neural Networks \cite{johnson2017inferring, hu2017learning, yi2018neural} are remarkably adept at answering questions about images in a task commonly known as  VQA \cite{antol2015vqa}. Until recently many of these methods were limited in their ability to generalize to other domains due to the difficulty of generating interfaces between modules. In \cite{suris2023vipergpt} this was addressed by using recent code generation techniques \cite{madaan2022language} from LLMs to create these interface layers. Specifically, these modular frameworks enable the generation of code which dictates the interaction between robust object detectors \cite{li2022grounded, li2023blip, he2017mask, redmon2016you, ge2021yolox}, depth estimators \cite{ranftl2020towards, zhao2020monocular, bhat2021adabins}. What's more remarkable is that these LLMs can also utilize existing functions inside of the Python language such as sorting and conditionals without any additional training \cite{vaithilingam2022expectation, nijkamp2022codegen, xu2022systematic, touvron2023llama}. Based on these exceptional findings we aim to extend the use of these modular concept-learning style frameworks to the robotic navigation domain.

\textbf{Path Planning for Robotic Navigation:} Path planning for robotic navigation and exploration is a long-studied research area \cite{mac2016heuristic, zhang2018path} with solutions ranging from sampling-based methods\cite{lavalle1998rapidly, karaman2011anytime} to graph-based lattice structures \cite{pivtoraiko2009differentially}. Variations of these probabilistic approaches are being used for self-driving cars \cite{lan2015continuous}, exploration \cite{ahmad2022efficient, biggie2023flexible, agha2021nebula, tranzatto2022cerberus} in complex environments. These approaches benefit from probabilistic guarantees on convergence and optimality \cite{lavalle1998rapidly} and adaptability to new environments making them natural choices to utilize in language grounding framework.

\textbf{Embodied Language-Based Navigation:} Recently a few other works have leveraged foundational models for embodied navigation. Clip on Wheels (CoW) is proposed in \cite{gadre2022clip} which combines text and image caption models \cite{radford2021learning} with frontier-based  exploration algorithms \cite{yamauchi1997frontier} to perform object exploration in simulation. In \cite{shah2023lm} landmark-based navigation is performed using pre-trained visual \cite{gadre2022clip}, language, and navigation modules \cite{shah2021ving}. Visual Language Navigation Maps (VLMaps) \cite{huang2022visual} learn a spatial language representation by combining RGB video feeds, code generation, and foundational models. SayCan \cite{ahn2022can}, and Palm-E \cite{driess2023palm} utilize the PALM \cite{chowdhery2022palm} language model to perform skill-based navigation and end-to-end navigation respectively. In general these methods are not tested on contextual examples, and some require the preprocessing of maps; in contrast, we test on situated examples with an experimental platform and require no situated priors.



\begin{figure}
    \centering
    \includegraphics[width=1\textwidth]{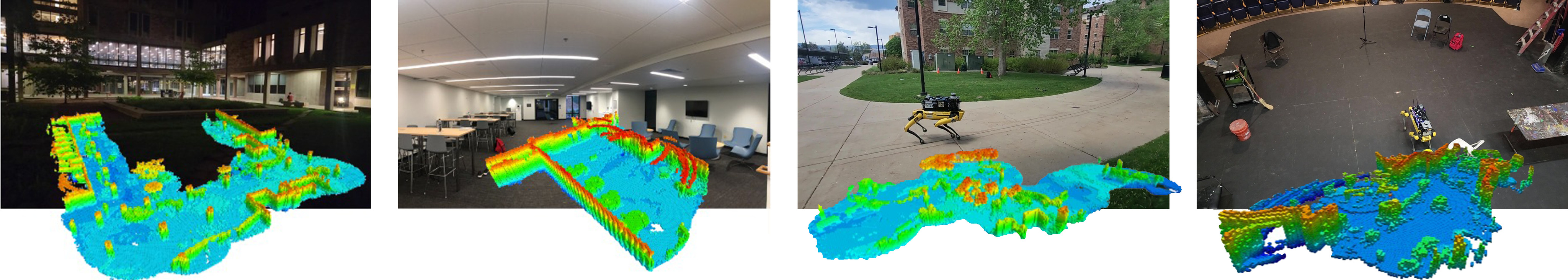}
    \begin{tabularx}{1.0\textwidth}{>{\centering\arraybackslash}X >{\centering\arraybackslash}X >{\centering\arraybackslash}X >{\centering\arraybackslash}X}
        (a) Courtyard & (b) Lobby & (c) Outdoor  & (d) Theater
    \end{tabularx}
    \caption[]{Full maps and representative images from the four testing environments.}
    \label{fig:environments}
\end{figure}

\section{Methodology}
\label{methodology}

\begin{figure}[htb]
    \centering
    \includegraphics[width=\textwidth]{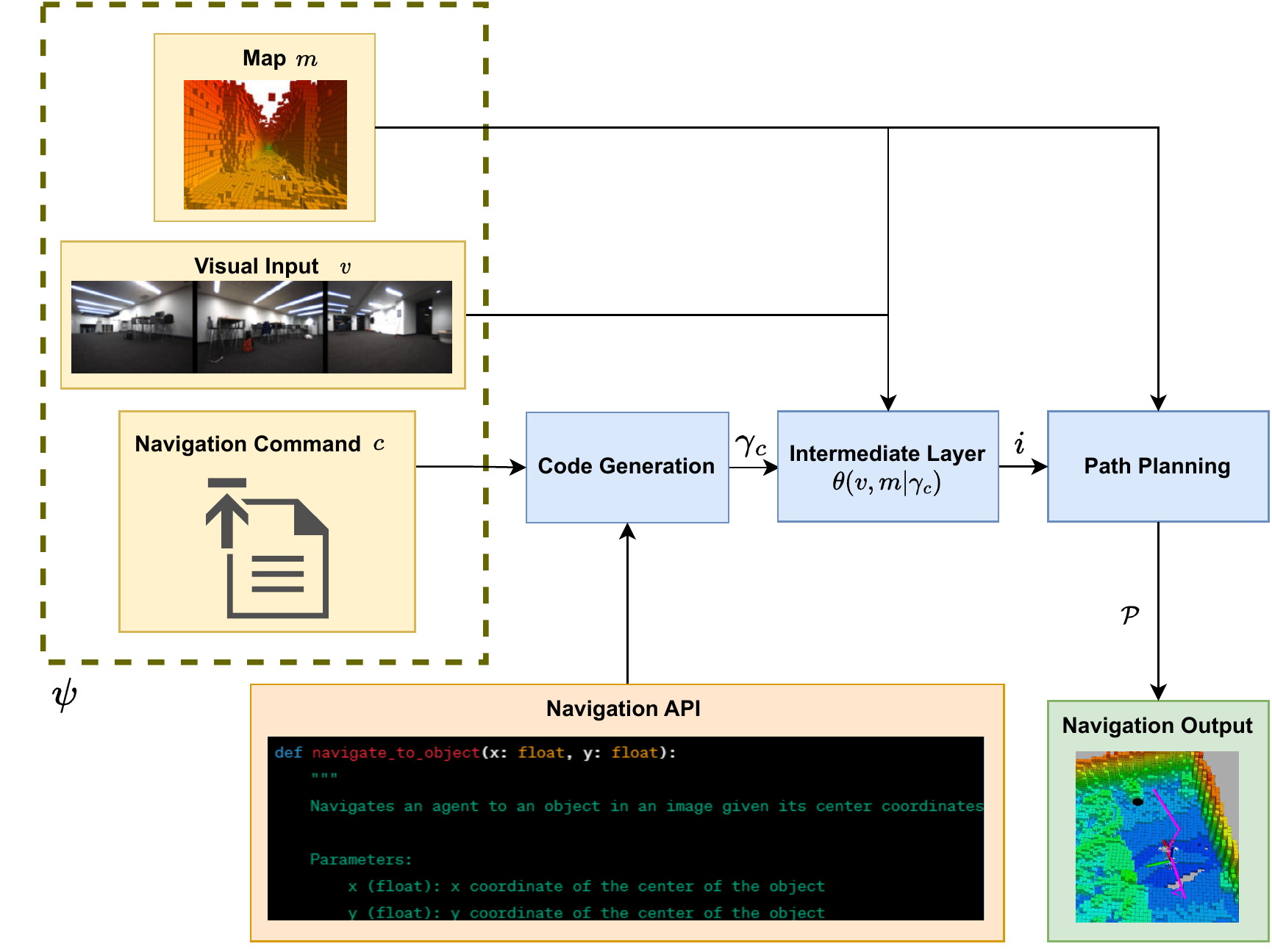}
    \caption{A graphical overview of the full system framework.}
    \label{fig:system}
\end{figure}

We develop a Navigation with Context Framework (NavCon) that leverages the rich contextual priors of LLMs and creates an executable code layer that interfaces with planning algorithms running on an embodied agent. Drawing on paradigms from modular neural networks \cite{andreas2016neural, gupta2023visual, suris2023vipergpt}, and concept learning techniques \cite{mao2019neuro}, we define a modular system that takes inputs at various stages in order to fully define a final navigational output for our system. This enables the use of arbitrary intermediate and substitutable layers which can also be executed on a distributed computing infrastructure. Formally, we have a collection of inputs $\psi$ composed of a visual input $v \in V$ an RGB image or multi-perspective collection of images, $m \in M$ a volumetric map of the space to be navigated, and $c \in C$ a natural language navigation command. Our system takes initially as input a command $c$ to generate code $\gamma_c$. This code consumes $v$ in order to resolve the grounding problem, i.e., to determine which object is being referred to in $c$ for navigational instructions, as well as $m$ in order to emplace the object into the world around the robot. This outputs $i = \theta(v,m|\gamma_c) \in \mathbb{R}^3$, a 3D waypoint to which navigation will commence in order to output $\mathcal{P}$ a robot trajectory which is a continuous function in $\mathbb{R}^3$. A graphical overview of the framework can be found in Figure \ref{fig:system}.

\textbf{Input Representations:} Our  visual input $v$ is in the form of an RGB image, either a semi-panoramic view or three separately labeled spatial images (left, front, right). Typical inference methods over RGB images \cite{redmon2016you, he2017mask, li2022grounded} reason in 2D over the image. For embodied navigation, we need to reason in 3D and associate different camera viewpoints with their associated 3D spatial relations. Embodied navigation introduces spatial relations that are difficult to reason over in 2D such as ``behind'', ``in front of'', or ``on your right''. To determine an appropriate input representation we evaluate the difference between sending in a concatenated image of all viewpoints (semi-panoramic) from the agent or sending in each frame separately with a spatial definition, i.e., right, front, or left. We find that spatial reasoning is best performed on a concatenated image and we present the details in Section \ref{sec:result}.

\textbf{Intermediate Layer:} We generate $\theta$ using recent advances in code generation models \cite{chen2021evaluating, nijkamp2022codegen}. We provide a functional set of navigation instructions in the form of a Python API for the code generation model. The full prompt containing the API can be found in the Appendix. These instructions include specifications for how to perform visual inference on the input $v$ using a similar paradigm found in \cite{suris2023vipergpt}. These API instructions include directions for finding an object, checking if an object has a certain property, etc. Additionally, we provide an API specification for interfacing with our geometric planner, resulting in code generated based on the natural language prompt $\gamma_c$.

\textbf{Planning:} We use a graph-based planner first presented in \cite{ahmad2022efficient, biggie2023flexible} where we provide a navigational endpoint. Specifically, we create a function that takes in the center coordinate $p$ of an object in image space and projects it as a landmark $l$ on a 3D map. The planner uses a sample-and-project approach similar to the one described in \cite{krusi2017driving, ahmad2022efficient} for terrain avoidance. Using this strategy, a graph of plans $\mathcal{G}$ is constructed by sampling points parameterized by the robot's width and length \cite{lavalle1998rapidly}. When a 3D waypoint is input to the planner we select the best path $\mathcal{P}$ from $\mathcal{G}$, for the robot to follow. If no path exists, we plan to the boundary of the graph and resample until the goal is reached. 

Waypoints are passed to the planner using the output $i$ from the intermediate layer. Specifically, image coordinates are translated into waypoints by executing $\theta$ and associating the result onto a 3D map $m$ \cite{hornung2013octomap} using ray casting. The map $m$ is generated online using \cite{shan2020lio} and the translation layer $\theta$ creates the necessary code to translate between the inputs $c$, $v$, $m$ as and the output $i$. We then plan to the waypoint by selecting the best path from $\mathcal{P}$.



\section{Experimental Results}
\label{sec:result}

We run two sets of experiments on a Boston Dynamics Spot equipped with a custom sensing suite consisting of a 3D 64 beam Ouster lidar, an imu, and 3 RGB cameras providing a semi-panoramic view of the environment. Our first experiment is designed to determine the best input representation for the visual layer and the second set tests the ability of our system to perform navigation in a wide variety of real-world environments.

For all of our experiments, we leverage ideas from human concept learning as in \cite{mao2019neuro} to categorize our sentences into four categories: Generic, Specific, Relational, and Contextual. Generic sentences are simply sentences that imply  ``Go to something'', e.g. ``Walk to the backpack.''. Specific sentences include a distinguishing piece of language such as a color attribute that directs the robot to one specific object in the scene. For example, if there are two backpacks, one red, and one black, then a specific sentence would be ``Drive to the black backpack.'' Relational sentences are any sentences that describe spatial relationships between objects in the scene such as, to the right of, on top of, etc. An example in this category is ``Move to the backpack on top of the chair.'' Contextual examples, require the robot to interpret the navigational goal based on background information. For instance, the sentence ``Find something that can carry water.'' requires the robot to know that a cup or a bucket can hold water. Our full list of sentences can be found in the Appendix.

\begin{figure}[htb]
    \centering
    \includegraphics[width=1\textwidth]{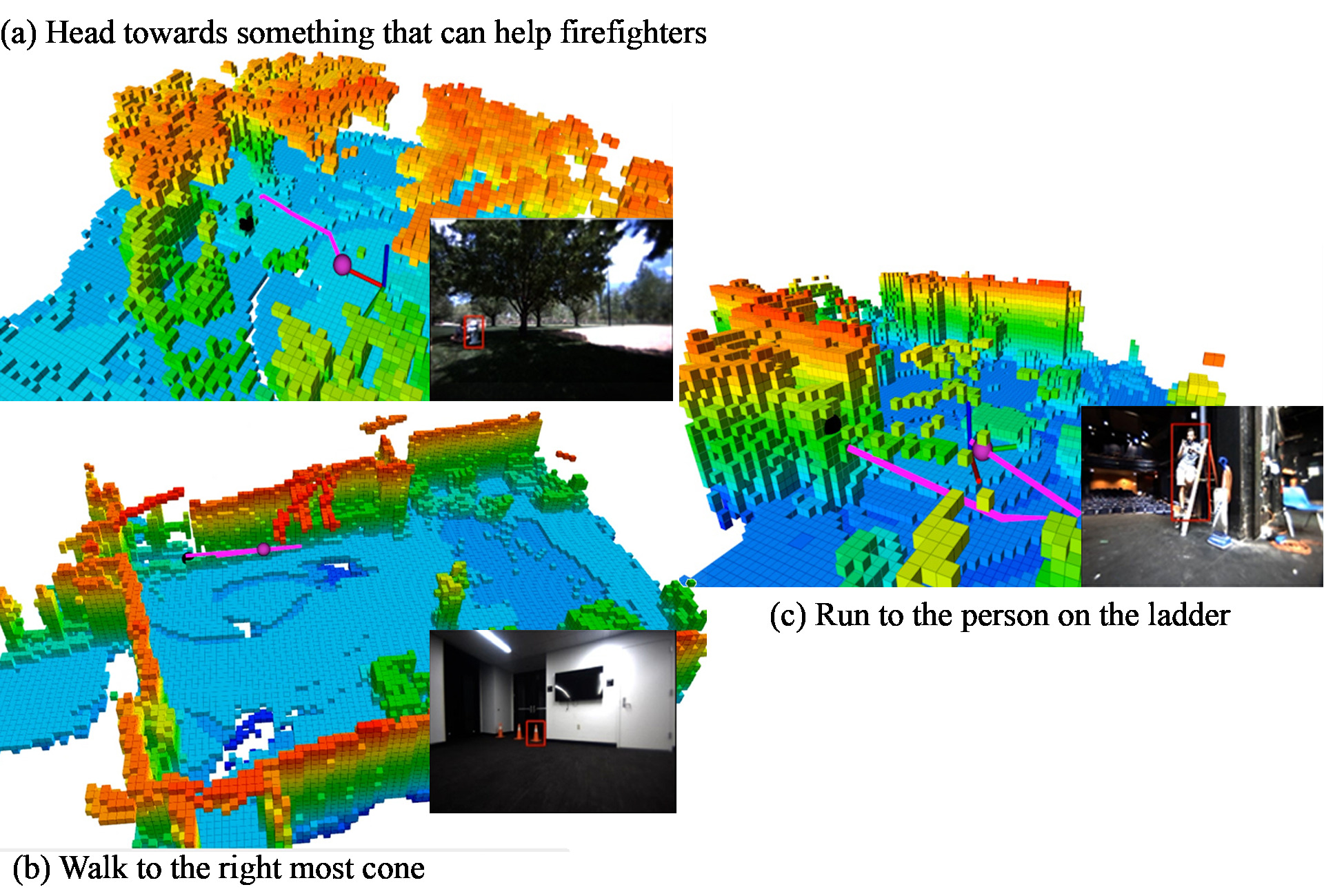}
    \caption{Example sentences, visual detections, maps, and planned paths.}
    \label{fig:sentence_example}
\end{figure}

Encoding spatial relationships into a model which has no concept of the physical world presents significant challenges. We determine an effective scheme for inputting three different viewpoints into the framework through empirical evaluations. Take the sentence, ''go to the chair on the right'' implies only the right half of the total FOV should be investigated but ''Go to the backpack that is to the right of the chair.'' could be in any image. To solve these limitations we use two different sets of input representations, the first is all three images stitched together with padding in between frames (\textbf{A}). In this case, we instruct the LLM on the order of the images (left, front, right) and let the model handle the spatial reasoning. In the second case, we process the frames individually (\textbf{B}) and let the model decide which frames to look at. We explicitly prompt the model with example code snippets that solve the difference between ``to your right'' and ``right of'' styles of language. at configuration. In this set of experiments, we evaluate the correctness of the generated code as well as if the code produced the correct object in the map $m$. This scene was created in a classroom setting with good lighting conditions as the emphasis was to evaluate the effect of the input representation. Results for successful code generation and object detections for the two input schemas are summarized in Table \ref{tab:classroom_results}.

\taburulecolor{gray!50}
\begin{table}[htb]
    \centering
    \rowcolors{2}{white}{gray!30}
    \begin{tabular}{|p{1.5cm}||p{1.5cm}|p{1.5cm}|p{1.5cm}|}
        \hline
        \textbf{Category} & \textbf{Count} & \textbf{A} (\%) & \textbf{B} (\%) \\
        \hline \hline
        Generic & 12 & 100 & 100 \\
        \hline
        Specific & 12 & 91.67 & 66.67 \\
        \hline
        Relational & 15 & 86.67 & 53.33 \\
        \hline
        Contextual & 11 & 81.82 & 45.45 \\
        \hline		
        Total & 50 & \textbf{90} & 66 \\
        \hline
    \end{tabular}
        \caption{Results showing the difference between the concatenated input representation \textbf{A} and the sequential representation \textbf{B}.}
        \label{tab:classroom_results}
\end{table}

From Table \ref{tab:classroom_results} we find that concatenating images significantly outperforms sending in the three frames separately. As expected, in the generic navigation case we are able to achieve 100\% success in generating the intermediate layer and for object identification. We see that the individual frame configuration (\textbf{B}) begins to break down when reasoning over specific objects. Specifically, the model fails to reason over spatial relations and object ordering when objects appear in more than one frame. For example, taking the sentence ``Go to the middle outlet'' only works when the three outlets are present in the same camera frame. If the middle outlet is in the front frame but the right outlet is in the right camera frame, this method fails. We explicitly see this in the generated code:

\begin{lstlisting}[language=Python, numbers=none]
outlet_patches.sort(key=lambda x: x.horizontal_center)
middle_outlet = outlet_patches[len(outlet_patches) // 2]
\end{lstlisting}

where each detected outlet is ordered based on its horizontal coordinate in the image. Since all outlets are merged into the same list the images have overlapping coordinates causing these failures. Configuration \textbf{B} struggles with relational sentences between objects for similar reasons.

In our second set of experiments, we evaluate the ability of our framework to navigate in real-world environments. The four environments shown in Figure \ref{fig:environments} range from an indoor lobby setting to a dark outdoor courtyard. We select these environments to highlight our framework's ability to generalize in multiple scenarios. In each environment, we test multiple sentences across the four categories (see the Appendix for the full list of sentences). Sentences are primarily based on landmarks already present in the environment but in some cases, we add additional artifacts to enable more extensive testing.

\taburulecolor{gray!50}
\begin{table}[htb]
    \centering
    \rowcolors{2}{white}{gray!30}
    \begin{tabular}{|p{1.7cm}||p{1.7cm}|p{1.7cm}|p{1.7cm}|p{1.7cm}|p{2.1cm}|}
        \hline
        \textbf{Category} & \textbf{Count} & \textbf{Code(\%)} & \textbf{OD(\%)} & \textbf{WP(\%)} & \textbf{Path\&Exec(\%)} \\
        \hline \hline
        Generic & 22 & 100 & 81.82 & 68.18 & 68.18 \\
        \hline
        Specific & 19 & 89.47 & 89.47 & 78.95 & 73.68 \\
        \hline
        Relational & 44 & 70.45 & 56.82 & 56.82 & 56.82 \\
        \hline
        Contextual & 29 & 65.52 & 41.38 & 41.38 & 41.38 \\
        \hline
        Total & 114 & 78.07 & 63.16 & 58.77 & 57.89 \\
        \hline
    \end{tabular}
    \caption{Summary of category-wise success rate for each of the four command types.}
    \label{tab:category_results}
\end{table}

We find that we are able to successfully generate code for navigational plans using a variety of verbs e.g., walk, go, drive, run, etc by leveraging the rich vocabulary knowledge present in foundational models. In fact, we can even say phrases like ``shashay to the stop sign'' which will be interpreted as a ``go to'' style navigation command.

\taburulecolor{gray!50}
\begin{table}[htb]
    \rowcolors{2}{white}{gray!30}
    \begin{tabular}{|p{1.7cm}||p{1.7cm}|p{1.7cm}|p{1.7cm}|p{1.7cm}|p{2.1cm}|}
        \hline
        \textbf{Scenes} & \textbf{Count} & \textbf{Code(\%)} & \textbf{OD(\%)} & \textbf{WP(\%)} & \textbf{Path\&Exec(\%)} \\
        \hline \hline
        Theater & 30 & 90 & 70 & 66.67 & 63.33 \\
        \hline
        Lobby & 29 & 65.52 & 48.28 & 44.83 & 44.83 \\
        \hline
        Outdoor & 24 & 87.5 & 79.17 & 70.83 & 70.83 \\
        \hline
        Courtyard & 31 & 70.97 & 58.06 & 54.84 & 54.84 \\
        \hline
        Total & 114 & 78.07 & 63.16 & 58.77 & 57.89 \\
        \hline
    \end{tabular}
        \caption{Summary of scene-wise success rate for each of the four command types. }
        \label{tab:scene_results}
\end{table}

From Tables \ref{tab:category_results} and \ref{tab:scene_results}, we see that for generic objects our framework has a 100\% success rate for code generation. In both tables \emph{Code} represents the success of the code generation step, \emph{OD} represents the success of the object detector (GLIP), \emph{WP} is the success of the 3D waypoint projection and \emph{Path\&Exec} is the success of the graph-based planning and navigation module. Our failures occurred in the object detection module where either the wrong object was detected or the object was not detected at all. However, since our framework is modular, the object detector can be substituted and as state-of-the-art vision detectors improve our framework will improve in turn. Additional failures occur in the waypoint projection step. These occurred for two reasons. In the first case, the vision system detected further out than the map resulting in projections at the edge of the map rather than the actual object. For the second case, projections missed the correct voxel on the map. This was either due to a ray ``clipping'' a closer object in the projection process. For longer projections (10--12m), the camera-to-lidar calibration caused enough error to make the projection inaccurate, which could be resolved through mean point clustering. We also note that we observed a 100\% success rate in following all planned paths.

\begin{figure}[htb]
    \centering
    \includegraphics[width=\textwidth]{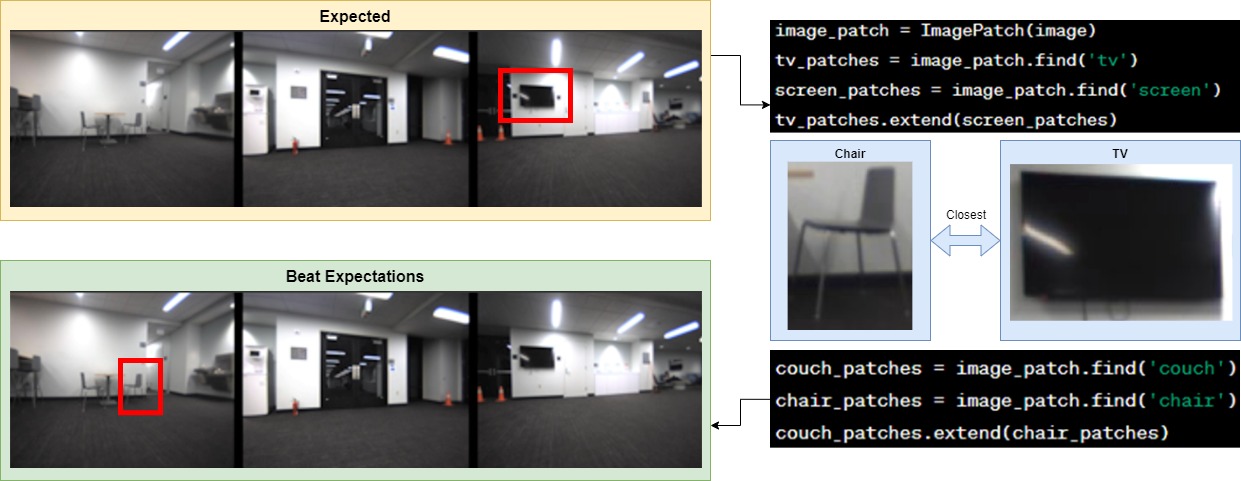}
    \caption{Results for the sentence ``Find me somewhere to watch a movie.''}
    \label{fig:tv}
\end{figure}

\textbf{Selected Qualitative Examples:}

Leveraging the vast knowledge base encoded into LLMs enables a degree of spatial reasoning and the ability to infer navigation instructions based on the context of the sentence. For example, take the sentence ``Find me something to help a firefighter'' as shown in Figure \ref{fig:firefighter} requires the robot to identify a list of objects that could help a firefighter and then look for them. From the snippet of the generated code, we see that the robot successfully looks for fire extinguishers and fire hydrants, both of which would aid a firefighter. Additionally, we are able to ask the robot to ``Find something to clean up a mess'' and it will find a mop or a broom.

What's more remarkable than simply interpreting correct navigation commands from context is our framework's ability to leverage foundational models to find better solutions to a task than we as humans would initially try and find. Take for example the sentence ``Find me somewhere to watch a movie,'' we would expect the robot to go and find a tv or some kind of screen. Of course, the first object the robot looks for is the screen but more remarkably the robot also looks for a chair which is closest to the screen.  We can see this in Figure \ref{fig:tv} where the generated layer $\theta$ explicitly looks for ``TVs, screens'' and then looks for ``chairs'' and ``couches.''

\begin{figure}[htb]
    \centering
    \includegraphics[width=0.8\textwidth]{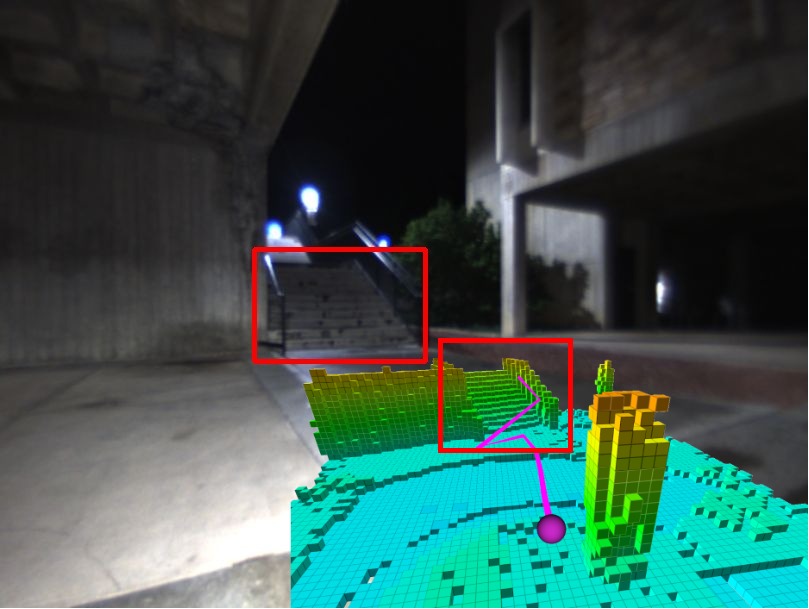}
    \caption{Scene used to tell the robot to ``Run upstairs'' and ``Go to the second floor''. The second sentence fails due to the code trying to ``count floors''}
    \label{fig:stair}
\end{figure}

As powerful as the contextual engine (GPT-3.5) is in our framework, it still has limitations when interpreting the physical world. In Figure \ref{fig:stair} we tell the robot to ``Run upstairs.'' and to no surprise, it will successfully generate code and plan up the stairs. However, when given the utterance ``Go to the second floor.'' the following code is generated:
\vspace{3cm}

\begin{lstlisting}[language=Python, numbers=none]
floor_patches = ImagePatch(image).find('floor')
floor_patches.sort(key=lambda x: x.vertical_center)
if len(floor_patches) < 2:
    return {'function': 'None', 'error': 'Image does not contain at least two floors.'}
second_floor_patch = floor_patches[1]
\end{lstlisting}

This code is nonsensical to the actual task because it's a literal interpretation of the sentence. Counting floors has nothing to do with going to the second story of a building. Examples like these highlight the need for additional methods of encoding 3D spatial reasoning into natural language based embodied navigation frameworks.

\section{Conclusion and Limitations}
\label{sec:conclusion}

NavCon, our composable framework shows remarkable accuracy and performance in generating code that fits the navigation task. In controlled experiments, we achieve 90\% accuracy across 50 different sentences. When taken to the real world the, model still performs exceptionally well with zero additional input. This is highlighted by a 70\% execution accuracy in large outdoor environments.

While this work makes significant strides toward embodied navigation, we observe many opportunities for future effort. Namely, operating in only 2D space creates challenges for spatial relationships such as ``behind'' the robot or ``to the right of.'' Additionally, we are limited by the visual range of the robot: therefore 3D projections need to be exponentially more accurate for both the smaller and further away an object is in the given input image. Finally, large multimodal foundational models are promising, but promise uncertain runtime performance given their lack of availability and scalability. Future work includes extending this framework to include more navigational inputs such as unbounded exploration in a direction or, having the robot explore until a condition is met. We also wish to include a module that operates on 3D maps and a module that supports sequential instructions.


\clearpage
\acknowledgments{This work was supported by USDA-NIFA Award Number 2021-67021-33450 and NSF IIS-HCC Project Number 1764092.}


\bibliography{example}  

\begin{thebibliography}{61}
\providecommand{\natexlab}[1]{#1}
\providecommand{\url}[1]{\texttt{#1}}
\expandafter\ifx\csname urlstyle\endcsname\relax
  \providecommand{\doi}[1]{doi: #1}\else
  \providecommand{\doi}{doi: \begingroup \urlstyle{rm}\Url}\fi

\bibitem[McNamara et~al.(1989)McNamara, Hardy, and
  Hirtle]{mcnamara1989subjective}
T.~P. McNamara, J.~K. Hardy, and S.~C. Hirtle.
\newblock Subjective hierarchies in spatial memory.
\newblock \emph{Journal of Experimental Psychology: Learning, Memory, and
  Cognition}, 15\penalty0 (2):\penalty0 211, 1989.

\bibitem[DARPA(2022)]{DARPA2022}
DARPA.
\newblock {DARPA Subterranean Challenge}, 2022.
\newblock URL \url{https://www.darpa.mil/program/darpa-subterranean-challenge}.
\newblock https://www.darpa.mil/program/darpa-subterranean-challenge.

\bibitem[Agha et~al.(2021)Agha, Otsu, Morrell, Fan, Thakker,
  Santamaria-Navarro, Kim, Bouman, Lei, Edlund, et~al.]{agha2021nebula}
A.~Agha, K.~Otsu, B.~Morrell, D.~D. Fan, R.~Thakker, A.~Santamaria-Navarro,
  S.-K. Kim, A.~Bouman, X.~Lei, J.~Edlund, et~al.
\newblock Nebula: Quest for robotic autonomy in challenging environments; team
  costar at the darpa subterranean challenge.
\newblock \emph{arXiv preprint arXiv:2103.11470}, 2021.

\bibitem[Biggie et~al.(2023)Biggie, Rush, Riley, Ahmad, Ohradzansky, Harlow,
  Miles, Torres, McGuire, Frew, et~al.]{biggie2023flexible}
H.~Biggie, E.~R. Rush, D.~G. Riley, S.~Ahmad, M.~T. Ohradzansky, K.~Harlow,
  M.~J. Miles, D.~Torres, S.~McGuire, E.~W. Frew, et~al.
\newblock Flexible supervised autonomy for exploration in subterranean
  environments.
\newblock \emph{arXiv preprint arXiv:2301.00771}, 2023.

\bibitem[Tranzatto et~al.(2022)Tranzatto, Miki, Dharmadhikari, Bernreiter,
  Kulkarni, Mascarich, Andersson, Khattak, Hutter, Siegwart,
  et~al.]{tranzatto2022cerberus}
M.~Tranzatto, T.~Miki, M.~Dharmadhikari, L.~Bernreiter, M.~Kulkarni,
  F.~Mascarich, O.~Andersson, S.~Khattak, M.~Hutter, R.~Siegwart, et~al.
\newblock Cerberus in the darpa subterranean challenge.
\newblock \emph{Science Robotics}, 7\penalty0 (66):\penalty0 eabp9742, 2022.

\bibitem[Zhao et~al.(2023)Zhao, Zhou, Li, Tang, Wang, Hou, Min, Zhang, Zhang,
  Dong, et~al.]{zhao2023survey}
W.~X. Zhao, K.~Zhou, J.~Li, T.~Tang, X.~Wang, Y.~Hou, Y.~Min, B.~Zhang,
  J.~Zhang, Z.~Dong, et~al.
\newblock A survey of large language models.
\newblock \emph{arXiv preprint arXiv:2303.18223}, 2023.

\bibitem[Glaese et~al.(2022)Glaese, McAleese, Trebacz, Aslanides, Firoiu,
  Ewalds, Rauh, Weidinger, Chadwick, Thacker, et~al.]{glaese2022improving}
A.~Glaese, N.~McAleese, M.~Trebacz, J.~Aslanides, V.~Firoiu, T.~Ewalds,
  M.~Rauh, L.~Weidinger, M.~Chadwick, P.~Thacker, et~al.
\newblock Improving alignment of dialogue agents via targeted human judgements.
\newblock \emph{arXiv preprint arXiv:2209.14375}, 2022.

\bibitem[Tsimpoukelli et~al.(2021)Tsimpoukelli, Menick, Cabi, Eslami, Vinyals,
  and Hill]{tsimpoukelli2021multimodal}
M.~Tsimpoukelli, J.~L. Menick, S.~Cabi, S.~Eslami, O.~Vinyals, and F.~Hill.
\newblock Multimodal few-shot learning with frozen language models.
\newblock \emph{Advances in Neural Information Processing Systems},
  34:\penalty0 200--212, 2021.

\bibitem[Liu et~al.(2023)Liu, Yuan, Fu, Jiang, Hayashi, and Neubig]{liu2023pre}
P.~Liu, W.~Yuan, J.~Fu, Z.~Jiang, H.~Hayashi, and G.~Neubig.
\newblock Pre-train, prompt, and predict: A systematic survey of prompting
  methods in natural language processing.
\newblock \emph{ACM Computing Surveys}, 55\penalty0 (9):\penalty0 1--35, 2023.

\bibitem[Chen et~al.(2021)Chen, Tworek, Jun, Yuan, Pinto, Kaplan, Edwards,
  Burda, Joseph, Brockman, et~al.]{chen2021evaluating}
M.~Chen, J.~Tworek, H.~Jun, Q.~Yuan, H.~P. d.~O. Pinto, J.~Kaplan, H.~Edwards,
  Y.~Burda, N.~Joseph, G.~Brockman, et~al.
\newblock Evaluating large language models trained on code.
\newblock \emph{arXiv preprint arXiv:2107.03374}, 2021.

\bibitem[Li et~al.(2022)Li, Choi, Chung, Kushman, Schrittwieser, Leblond,
  Eccles, Keeling, Gimeno, Dal~Lago, et~al.]{li2022competition}
Y.~Li, D.~Choi, J.~Chung, N.~Kushman, J.~Schrittwieser, R.~Leblond, T.~Eccles,
  J.~Keeling, F.~Gimeno, A.~Dal~Lago, et~al.
\newblock Competition-level code generation with alphacode.
\newblock \emph{Science}, 378\penalty0 (6624):\penalty0 1092--1097, 2022.

\bibitem[Xu et~al.(2022)Xu, Alon, Neubig, and Hellendoorn]{xu2022systematic}
F.~F. Xu, U.~Alon, G.~Neubig, and V.~J. Hellendoorn.
\newblock A systematic evaluation of large language models of code.
\newblock In \emph{Proceedings of the 6th ACM SIGPLAN International Symposium
  on Machine Programming}, pages 1--10, 2022.

\bibitem[Alayrac et~al.(2022)Alayrac, Donahue, Luc, Miech, Barr, Hasson, Lenc,
  Mensch, Millican, Reynolds, et~al.]{alayrac2022flamingo}
J.-B. Alayrac, J.~Donahue, P.~Luc, A.~Miech, I.~Barr, Y.~Hasson, K.~Lenc,
  A.~Mensch, K.~Millican, M.~Reynolds, et~al.
\newblock Flamingo: a visual language model for few-shot learning.
\newblock \emph{Advances in Neural Information Processing Systems},
  35:\penalty0 23716--23736, 2022.

\bibitem[Sur{\'\i}s et~al.(2023)Sur{\'\i}s, Menon, and
  Vondrick]{suris2023vipergpt}
D.~Sur{\'\i}s, S.~Menon, and C.~Vondrick.
\newblock Vipergpt: Visual inference via python execution for reasoning.
\newblock \emph{arXiv preprint arXiv:2303.08128}, 2023.

\bibitem[Li et~al.(2022)Li, Zhang, Zhang, Yang, Li, Zhong, Wang, Yuan, Zhang,
  Hwang, et~al.]{li2022grounded}
L.~H. Li, P.~Zhang, H.~Zhang, J.~Yang, C.~Li, Y.~Zhong, L.~Wang, L.~Yuan,
  L.~Zhang, J.-N. Hwang, et~al.
\newblock Grounded language-image pre-training.
\newblock In \emph{Proceedings of the IEEE/CVF Conference on Computer Vision
  and Pattern Recognition}, pages 10965--10975, 2022.

\bibitem[LaValle et~al.(1998)]{lavalle1998rapidly}
S.~M. LaValle et~al.
\newblock Rapidly-exploring random trees: A new tool for path planning.
\newblock 1998.

\bibitem[Dang et~al.(2020)Dang, Tranzatto, Khattak, Mascarich, Alexis, and
  Hutter]{dang2020graph}
T.~Dang, M.~Tranzatto, S.~Khattak, F.~Mascarich, K.~Alexis, and M.~Hutter.
\newblock Graph-based subterranean exploration path planning using aerial and
  legged robots.
\newblock \emph{Journal of Field Robotics}, 37\penalty0 (8):\penalty0
  1363--1388, 2020.

\bibitem[Ahmad and Humbert(2022)]{ahmad2022efficient}
S.~Ahmad and J.~S. Humbert.
\newblock Efficient sampling-based planning for subterranean exploration.
\newblock In \emph{2022 IEEE/RSJ International Conference on Intelligent Robots
  and Systems (IROS)}, pages 7114--7121. IEEE, 2022.

\bibitem[Tellex et~al.(2011)Tellex, Kollar, Dickerson, Walter, Banerjee,
  Teller, and Roy]{tellex2011understanding}
S.~Tellex, T.~Kollar, S.~Dickerson, M.~Walter, A.~Banerjee, S.~Teller, and
  N.~Roy.
\newblock Understanding natural language commands for robotic navigation and
  mobile manipulation.
\newblock In \emph{Proceedings of the AAAI Conference on Artificial
  Intelligence}, volume~25, pages 1507--1514, 2011.

\bibitem[Howard et~al.(2014)Howard, Tellex, and Roy]{howard2014natural}
T.~M. Howard, S.~Tellex, and N.~Roy.
\newblock A natural language planner interface for mobile manipulators.
\newblock In \emph{2014 IEEE International Conference on Robotics and
  Automation (ICRA)}, pages 6652--6659. IEEE, 2014.

\bibitem[Kollar et~al.(2010)Kollar, Tellex, Roy, and Roy]{kollar2010toward}
T.~Kollar, S.~Tellex, D.~Roy, and N.~Roy.
\newblock Toward understanding natural language directions.
\newblock In \emph{2010 5th ACM/IEEE International Conference on Human-Robot
  Interaction (HRI)}, pages 259--266. IEEE, 2010.

\bibitem[Driess et~al.(2023)Driess, Xia, Sajjadi, Lynch, Chowdhery, Ichter,
  Wahid, Tompson, Vuong, Yu, et~al.]{driess2023palm}
D.~Driess, F.~Xia, M.~S. Sajjadi, C.~Lynch, A.~Chowdhery, B.~Ichter, A.~Wahid,
  J.~Tompson, Q.~Vuong, T.~Yu, et~al.
\newblock Palm-e: An embodied multimodal language model.
\newblock \emph{arXiv preprint arXiv:2303.03378}, 2023.

\bibitem[Floridi and Chiriatti(2020)]{floridi2020gpt}
L.~Floridi and M.~Chiriatti.
\newblock Gpt-3: Its nature, scope, limits, and consequences.
\newblock \emph{Minds and Machines}, 30:\penalty0 681--694, 2020.

\bibitem[Touvron et~al.(2023)Touvron, Lavril, Izacard, Martinet, Lachaux,
  Lacroix, Rozi{\`e}re, Goyal, Hambro, Azhar, et~al.]{touvron2023llama}
H.~Touvron, T.~Lavril, G.~Izacard, X.~Martinet, M.-A. Lachaux, T.~Lacroix,
  B.~Rozi{\`e}re, N.~Goyal, E.~Hambro, F.~Azhar, et~al.
\newblock Llama: Open and efficient foundation language models.
\newblock \emph{arXiv preprint arXiv:2302.13971}, 2023.

\bibitem[Vaswani et~al.(2017)Vaswani, Shazeer, Parmar, Uszkoreit, Jones, Gomez,
  Kaiser, and Polosukhin]{vaswani2017attention}
A.~Vaswani, N.~Shazeer, N.~Parmar, J.~Uszkoreit, L.~Jones, A.~N. Gomez,
  {\L}.~Kaiser, and I.~Polosukhin.
\newblock Attention is all you need.
\newblock \emph{Advances in neural information processing systems}, 30, 2017.

\bibitem[Brohan et~al.(2023)Brohan, Chebotar, Finn, Hausman, Herzog, Ho, Ibarz,
  Irpan, Jang, Julian, et~al.]{brohan2023can}
A.~Brohan, Y.~Chebotar, C.~Finn, K.~Hausman, A.~Herzog, D.~Ho, J.~Ibarz,
  A.~Irpan, E.~Jang, R.~Julian, et~al.
\newblock Do as i can, not as i say: Grounding language in robotic affordances.
\newblock In \emph{Conference on Robot Learning}, pages 287--318. PMLR, 2023.

\bibitem[Sajjadi et~al.(2022)Sajjadi, Duckworth, Mahendran, van Steenkiste,
  Pavetic, Lucic, Guibas, Greff, and Kipf]{sajjadi2022object}
M.~S. Sajjadi, D.~Duckworth, A.~Mahendran, S.~van Steenkiste, F.~Pavetic,
  M.~Lucic, L.~J. Guibas, K.~Greff, and T.~Kipf.
\newblock Object scene representation transformer.
\newblock \emph{Advances in Neural Information Processing Systems},
  35:\penalty0 9512--9524, 2022.

\bibitem[Noreen et~al.(2016)Noreen, Khan, and Habib]{noreen2016optimal}
I.~Noreen, A.~Khan, and Z.~Habib.
\newblock Optimal path planning using rrt* based approaches: a survey and
  future directions.
\newblock \emph{International Journal of Advanced Computer Science and
  Applications}, 7\penalty0 (11), 2016.

\bibitem[Johnson et~al.(2017)Johnson, Hariharan, Van Der~Maaten, Hoffman,
  Fei-Fei, Lawrence~Zitnick, and Girshick]{johnson2017inferring}
J.~Johnson, B.~Hariharan, L.~Van Der~Maaten, J.~Hoffman, L.~Fei-Fei,
  C.~Lawrence~Zitnick, and R.~Girshick.
\newblock Inferring and executing programs for visual reasoning.
\newblock In \emph{Proceedings of the IEEE international conference on computer
  vision}, pages 2989--2998, 2017.

\bibitem[Hu et~al.(2017)Hu, Andreas, Rohrbach, Darrell, and
  Saenko]{hu2017learning}
R.~Hu, J.~Andreas, M.~Rohrbach, T.~Darrell, and K.~Saenko.
\newblock Learning to reason: End-to-end module networks for visual question
  answering.
\newblock In \emph{Proceedings of the IEEE international conference on computer
  vision}, pages 804--813, 2017.

\bibitem[Yi et~al.(2018)Yi, Wu, Gan, Torralba, Kohli, and
  Tenenbaum]{yi2018neural}
K.~Yi, J.~Wu, C.~Gan, A.~Torralba, P.~Kohli, and J.~Tenenbaum.
\newblock Neural-symbolic vqa: Disentangling reasoning from vision and language
  understanding.
\newblock \emph{Advances in neural information processing systems}, 31, 2018.

\bibitem[Antol et~al.(2015)Antol, Agrawal, Lu, Mitchell, Batra, Zitnick, and
  Parikh]{antol2015vqa}
S.~Antol, A.~Agrawal, J.~Lu, M.~Mitchell, D.~Batra, C.~L. Zitnick, and
  D.~Parikh.
\newblock Vqa: Visual question answering.
\newblock In \emph{Proceedings of the IEEE international conference on computer
  vision}, pages 2425--2433, 2015.

\bibitem[Madaan et~al.(2022)Madaan, Zhou, Alon, Yang, and
  Neubig]{madaan2022language}
A.~Madaan, S.~Zhou, U.~Alon, Y.~Yang, and G.~Neubig.
\newblock Language models of code are few-shot commonsense learners.
\newblock \emph{arXiv preprint arXiv:2210.07128}, 2022.

\bibitem[Li et~al.(2023)Li, Li, Savarese, and Hoi]{li2023blip}
J.~Li, D.~Li, S.~Savarese, and S.~Hoi.
\newblock Blip-2: Bootstrapping language-image pre-training with frozen image
  encoders and large language models.
\newblock \emph{arXiv preprint arXiv:2301.12597}, 2023.

\bibitem[He et~al.(2017)He, Gkioxari, Doll{\'a}r, and Girshick]{he2017mask}
K.~He, G.~Gkioxari, P.~Doll{\'a}r, and R.~Girshick.
\newblock Mask r-cnn.
\newblock In \emph{Proceedings of the IEEE international conference on computer
  vision}, pages 2961--2969, 2017.

\bibitem[Redmon et~al.(2016)Redmon, Divvala, Girshick, and
  Farhadi]{redmon2016you}
J.~Redmon, S.~Divvala, R.~Girshick, and A.~Farhadi.
\newblock You only look once: Unified, real-time object detection.
\newblock In \emph{Proceedings of the IEEE conference on computer vision and
  pattern recognition}, pages 779--788, 2016.

\bibitem[Ge et~al.(2021)Ge, Liu, Wang, Li, and Sun]{ge2021yolox}
Z.~Ge, S.~Liu, F.~Wang, Z.~Li, and J.~Sun.
\newblock Yolox: Exceeding yolo series in 2021.
\newblock \emph{arXiv preprint arXiv:2107.08430}, 2021.

\bibitem[Ranftl et~al.(2020)Ranftl, Lasinger, Hafner, Schindler, and
  Koltun]{ranftl2020towards}
R.~Ranftl, K.~Lasinger, D.~Hafner, K.~Schindler, and V.~Koltun.
\newblock Towards robust monocular depth estimation: Mixing datasets for
  zero-shot cross-dataset transfer.
\newblock \emph{IEEE transactions on pattern analysis and machine
  intelligence}, 44\penalty0 (3):\penalty0 1623--1637, 2020.

\bibitem[Zhao et~al.(2020)Zhao, Sun, Zhang, Tang, and Qian]{zhao2020monocular}
C.~Zhao, Q.~Sun, C.~Zhang, Y.~Tang, and F.~Qian.
\newblock Monocular depth estimation based on deep learning: An overview.
\newblock \emph{Science China Technological Sciences}, 63\penalty0
  (9):\penalty0 1612--1627, 2020.

\bibitem[Bhat et~al.(2021)Bhat, Alhashim, and Wonka]{bhat2021adabins}
S.~F. Bhat, I.~Alhashim, and P.~Wonka.
\newblock Adabins: Depth estimation using adaptive bins.
\newblock In \emph{Proceedings of the IEEE/CVF Conference on Computer Vision
  and Pattern Recognition}, pages 4009--4018, 2021.

\bibitem[Vaithilingam et~al.(2022)Vaithilingam, Zhang, and
  Glassman]{vaithilingam2022expectation}
P.~Vaithilingam, T.~Zhang, and E.~L. Glassman.
\newblock Expectation vs. experience: Evaluating the usability of code
  generation tools powered by large language models.
\newblock In \emph{Chi conference on human factors in computing systems
  extended abstracts}, pages 1--7, 2022.

\bibitem[Nijkamp et~al.(2022)Nijkamp, Pang, Hayashi, Tu, Wang, Zhou, Savarese,
  and Xiong]{nijkamp2022codegen}
E.~Nijkamp, B.~Pang, H.~Hayashi, L.~Tu, H.~Wang, Y.~Zhou, S.~Savarese, and
  C.~Xiong.
\newblock Codegen: An open large language model for code with multi-turn
  program synthesis.
\newblock \emph{arXiv preprint arXiv:2203.13474}, 2022.

\bibitem[Mac et~al.(2016)Mac, Copot, Tran, and De~Keyser]{mac2016heuristic}
T.~T. Mac, C.~Copot, D.~T. Tran, and R.~De~Keyser.
\newblock Heuristic approaches in robot path planning: A survey.
\newblock \emph{Robotics and Autonomous Systems}, 86:\penalty0 13--28, 2016.

\bibitem[Zhang et~al.(2018)Zhang, Lin, and Chen]{zhang2018path}
H.-y. Zhang, W.-m. Lin, and A.-x. Chen.
\newblock Path planning for the mobile robot: A review.
\newblock \emph{Symmetry}, 10\penalty0 (10):\penalty0 450, 2018.

\bibitem[Karaman et~al.(2011)Karaman, Walter, Perez, Frazzoli, and
  Teller]{karaman2011anytime}
S.~Karaman, M.~R. Walter, A.~Perez, E.~Frazzoli, and S.~Teller.
\newblock Anytime motion planning using the rrt.
\newblock In \emph{2011 IEEE international conference on robotics and
  automation}, pages 1478--1483. IEEE, 2011.

\bibitem[Pivtoraiko et~al.(2009)Pivtoraiko, Knepper, and
  Kelly]{pivtoraiko2009differentially}
M.~Pivtoraiko, R.~A. Knepper, and A.~Kelly.
\newblock Differentially constrained mobile robot motion planning in state
  lattices.
\newblock \emph{Journal of Field Robotics}, 26\penalty0 (3):\penalty0 308--333,
  2009.

\bibitem[Lan and Di~Cairano(2015)]{lan2015continuous}
X.~Lan and S.~Di~Cairano.
\newblock Continuous curvature path planning for semi-autonomous vehicle
  maneuvers using rrt.
\newblock In \emph{2015 European control conference (ECC)}, pages 2360--2365.
  IEEE, 2015.

\bibitem[Gadre et~al.(2022)Gadre, Wortsman, Ilharco, Schmidt, and
  Song]{gadre2022clip}
S.~Y. Gadre, M.~Wortsman, G.~Ilharco, L.~Schmidt, and S.~Song.
\newblock Clip on wheels: Zero-shot object navigation as object localization
  and exploration.
\newblock \emph{arXiv preprint arXiv:2203.10421}, 2022.

\bibitem[Radford et~al.(2021)Radford, Kim, Hallacy, Ramesh, Goh, Agarwal,
  Sastry, Askell, Mishkin, Clark, et~al.]{radford2021learning}
A.~Radford, J.~W. Kim, C.~Hallacy, A.~Ramesh, G.~Goh, S.~Agarwal, G.~Sastry,
  A.~Askell, P.~Mishkin, J.~Clark, et~al.
\newblock Learning transferable visual models from natural language
  supervision.
\newblock In \emph{International conference on machine learning}, pages
  8748--8763. PMLR, 2021.

\bibitem[Yamauchi(1997)]{yamauchi1997frontier}
B.~Yamauchi.
\newblock A frontier-based approach for autonomous exploration.
\newblock In \emph{Proceedings 1997 IEEE International Symposium on
  Computational Intelligence in Robotics and Automation CIRA'97.'Towards New
  Computational Principles for Robotics and Automation'}, pages 146--151. IEEE,
  1997.

\bibitem[Shah et~al.(2023)Shah, Osi{\'n}ski, Levine, et~al.]{shah2023lm}
D.~Shah, B.~Osi{\'n}ski, S.~Levine, et~al.
\newblock Lm-nav: Robotic navigation with large pre-trained models of language,
  vision, and action.
\newblock In \emph{Conference on Robot Learning}, pages 492--504. PMLR, 2023.

\bibitem[Shah et~al.(2021)Shah, Eysenbach, Kahn, Rhinehart, and
  Levine]{shah2021ving}
D.~Shah, B.~Eysenbach, G.~Kahn, N.~Rhinehart, and S.~Levine.
\newblock Ving: Learning open-world navigation with visual goals.
\newblock In \emph{2021 IEEE International Conference on Robotics and
  Automation (ICRA)}, pages 13215--13222. IEEE, 2021.

\bibitem[Huang et~al.(2022)Huang, Mees, Zeng, and Burgard]{huang2022visual}
C.~Huang, O.~Mees, A.~Zeng, and W.~Burgard.
\newblock Visual language maps for robot navigation.
\newblock \emph{arXiv preprint arXiv:2210.05714}, 2022.

\bibitem[Ahn et~al.(2022)Ahn, Brohan, Brown, Chebotar, Cortes, David, Finn,
  Gopalakrishnan, Hausman, Herzog, et~al.]{ahn2022can}
M.~Ahn, A.~Brohan, N.~Brown, Y.~Chebotar, O.~Cortes, B.~David, C.~Finn,
  K.~Gopalakrishnan, K.~Hausman, A.~Herzog, et~al.
\newblock Do as i can, not as i say: Grounding language in robotic affordances.
\newblock \emph{arXiv preprint arXiv:2204.01691}, 2022.

\bibitem[Chowdhery et~al.(2022)Chowdhery, Narang, Devlin, Bosma, Mishra,
  Roberts, Barham, Chung, Sutton, Gehrmann, et~al.]{chowdhery2022palm}
A.~Chowdhery, S.~Narang, J.~Devlin, M.~Bosma, G.~Mishra, A.~Roberts, P.~Barham,
  H.~W. Chung, C.~Sutton, S.~Gehrmann, et~al.
\newblock Palm: Scaling language modeling with pathways.
\newblock \emph{arXiv preprint arXiv:2204.02311}, 2022.

\bibitem[Andreas et~al.(2016)Andreas, Rohrbach, Darrell, and
  Klein]{andreas2016neural}
J.~Andreas, M.~Rohrbach, T.~Darrell, and D.~Klein.
\newblock Neural module networks.
\newblock In \emph{Proceedings of the IEEE conference on computer vision and
  pattern recognition}, pages 39--48, 2016.

\bibitem[Gupta and Kembhavi(2023)]{gupta2023visual}
T.~Gupta and A.~Kembhavi.
\newblock Visual programming: Compositional visual reasoning without training.
\newblock In \emph{Proceedings of the IEEE/CVF Conference on Computer Vision
  and Pattern Recognition}, pages 14953--14962, 2023.

\bibitem[Mao et~al.(2019)Mao, Gan, Kohli, Tenenbaum, and Wu]{mao2019neuro}
J.~Mao, C.~Gan, P.~Kohli, J.~B. Tenenbaum, and J.~Wu.
\newblock The neuro-symbolic concept learner: Interpreting scenes, words, and
  sentences from natural supervision.
\newblock \emph{arXiv preprint arXiv:1904.12584}, 2019.

\bibitem[Kr{\"u}si et~al.(2017)Kr{\"u}si, Furgale, Bosse, and
  Siegwart]{krusi2017driving}
P.~Kr{\"u}si, P.~Furgale, M.~Bosse, and R.~Siegwart.
\newblock Driving on point clouds: Motion planning, trajectory optimization,
  and terrain assessment in generic nonplanar environments.
\newblock \emph{Journal of Field Robotics}, 34\penalty0 (5):\penalty0 940--984,
  2017.

\bibitem[Hornung et~al.(2013)Hornung, Wurm, Bennewitz, Stachniss, and
  Burgard]{hornung2013octomap}
A.~Hornung, K.~M. Wurm, M.~Bennewitz, C.~Stachniss, and W.~Burgard.
\newblock Octomap: An efficient probabilistic 3d mapping framework based on
  octrees.
\newblock \emph{Autonomous robots}, 34:\penalty0 189--206, 2013.

\bibitem[Shan et~al.(2020)Shan, Englot, Meyers, Wang, Ratti, and
  Rus]{shan2020lio}
T.~Shan, B.~Englot, D.~Meyers, W.~Wang, C.~Ratti, and D.~Rus.
\newblock Lio-sam: Tightly-coupled lidar inertial odometry via smoothing and
  mapping.
\newblock In \emph{2020 IEEE/RSJ international conference on intelligent robots
  and systems (IROS)}, pages 5135--5142. IEEE, 2020.

\end{thebibliography}

\newpage
\section{Appendix}
\subsection{Navigation Prompts}

The framework is tested on 4 different real-world environments. Further details about each of the environments are listed below:
\begin{itemize}
    \item {\textbf{Theater}, an indoor environment simulating a stage-like environment during set construction.} 
    \item {\textbf{Lobby}, an indoor environment, mostly occupied with chairs and tables, posing more challenges in distinguishing specific objects and planning}
    \item {\textbf{Outdoor}, a larger environment with objects such as trees, cars, bikes, and sporting areas.}
    \item {\textbf{Courtyard}, an outdoor environment tested at night to challenge the framework with low-light images.}
\end{itemize}
Along with the general objects present in the scene, we placed some additional items like backpacks, shoes, and cones into the scene. All the prompts tested in these 4 different environments are presented in Tables \ref{theater_sentences}, \ref{lobby_sentences}, \ref{outdoor_sentences}, and \ref{courtyard_sentences}. For each of the prompts, we also highlighted whether each step succeeded or not. 
\taburulecolor{gray!50}
\begin{table}
    \caption{Scene-01: Theater}
    \label{theater_sentences}
    \rowcolors{2}{white}{gray!30}
    \begin{tabular}{|p{1.5cm}|p{7cm}|p{0.9cm}|p{0.9cm}|p{0.9cm}|p{0.9cm}|}
        \hline
        \textbf{Category} & \textbf{Scene} & \textbf{Code} & \textbf{OD} & \textbf{WP} & \textbf{Path\& Exec} \\
        \hline \hline
        Generic & Drive to the helmet	& Pass & Pass & Pass & Pass \\
        \hline
        Generic & Navigate to the mop	& Pass & Pass & Pass & Pass \\
        \hline
        Generic & Go to the vacuum & Pass & Fail & Fail & Fail \\
        \hline
        Generic & Walk to the table & Pass & Pass & Pass & Pass \\
        \hline
        Generic & Go to the fire extinguisher	& Pass & Pass & Pass & Pass \\
        \hline
        Generic & Walk to the saw	& Pass & Fail & Fail & Fail \\
        \hline
        Specific & Go to the red backpack & Pass & Pass & Pass & Pass \\
        \hline
        Specific & Move to the orange bucket & Pass & Pass & Pass & Pass \\
        \hline
        Specific & Drive to the chair on the right	& Pass & Pass & Pass & Fail \\
        \hline
        Specific & Walk to the white fan & Pass & Pass & Pass & Pass \\
        \hline
        Specific & Navigate to the black fan & Pass & Pass & Pass & Pass \\
        \hline
        Specific & Navigate to the black speaker & Pass & Pass & Pass & Pass \\
        \hline
        Specific & Run to the red bag & Pass & Pass & Fail & Fail \\
        \hline
        Specific & Move to the black chair infront of you & Pass & Pass & Pass & Pass \\
        \hline
        Specific & Walk to the blue chair & Pass & Pass & Pass & Pass \\
        \hline
        Relational & Go to the chair with the black backpack on it & Pass & Pass & Pass & Pass \\
        \hline
        Relational & Go to the chair with the helmet on it & Pass & Pass & Pass & Pass \\
        \hline
        Relational & Go to the backpack next to the chair & Pass & Fail & Fail & Fail \\
        \hline
        Relational & Go to the backpack infront of the ladder & Pass & Pass & Pass & Pass \\
        \hline
        Relational & Run to the person on the ladder	& Pass & Pass & Pass & Pass \\
        \hline
        Relational & Drive to the man sitting on the table & Pass & Pass & Pass & Pass \\
        \hline
        Relational & Go to the third chair from the left	& Pass & Pass & Pass & Pass \\
        \hline
        Contextual & Go to something that can carry water & Pass & Pass & Pass & Pass \\
        \hline
        Contextual & Navigate to the somethings that can wash the floor & Fail & Fail & Fail & Fail \\
        \hline
        Contextual & My floor is dirty. Go to something that can fix it & Fail & Fail & Fail & Fail \\
        \hline
        Contextual & Go somewhere that I throw something in & Pass & Pass & Pass & Pass \\
        \hline
        Contextual & Walk to something that put out the fire	& Pass & Fail & Fail & Fail \\
        \hline
        Contextual & Go to something that can cool me down & Pass & Fail & Fail & Fail \\
        \hline
        Contextual & Go to something white that can cool me down	& Fail & Fail & Fail & Fail \\
        \hline
        Contextual & I am in the mood to listen to music. Go to something that can do that & Pass & Fail & Fail & Fail \\
        \hline
    \end{tabular}
\end{table}

\begin{table}
    \caption{Scene-02: Lobby}
    \label{lobby_sentences}
    \rowcolors{2}{white}{gray!30}
    \begin{tabular}{|p{1.5cm}|p{7cm}|p{0.9cm}|p{0.9cm}|p{0.9cm}|p{0.9cm}|}
        \hline
        \textbf{Category} & \textbf{Scene} & \textbf{Code} & \textbf{OD} & \textbf{WP} & \textbf{Path\& Exec} \\
        \hline \hline
        Generic & Run to the trash bag & Pass & Pass & Pass & Pass \\
        \hline
        Generic & Navigate to the water fountain & Pass & Fail & Fail & Fail \\
        \hline
        Generic & Drive to the broom & Pass & Pass & Pass & Pass \\
        \hline
        Generic & Walk to monitor & Pass & Pass & Fail & Fail \\
        \hline
        Specific & Walk to the tiny monitor & Fail & Fail & Fail & Fail \\
        \hline
        Specific & Walk to the smallest monitor & Pass & Pass & Pass & Pass \\
        \hline
        Relational & Walk to the right most cone & Pass & Pass & Pass & Pass \\
        \hline
        Relational & Walk to the table with the can on it & Pass & Fail & Fail & Fail \\
        \hline
        Relational & Walk to the table next to the red backpack & Pass & Pass & Pass & Pass \\
        \hline
        Relational & Go the blue chair with the backpack on it & Pass & Pass & Pass & Pass \\
        \hline
        Relational & Walk to the leftmost table & Pass & Fail & Fail & Fail \\
        \hline
        Relational & Go to backpack closest to the shoe & Pass & Pass & Pass & Pass \\
        \hline
        Relational & Walk to shoe next to the red backpack & Pass & Pass & Pass & Pass \\
        \hline
        Relational & Remove the trash from the floor & Fail & Fail & Fail & Fail \\
        \hline
        Relational & Walk to the table with the monitor & Fail & Fail & Fail & Fail \\
        \hline
        Relational & Drive to the closest monitor to the table & Pass & Pass & Pass & Pass \\
        \hline
        Relational & Go to the table with the bottle on it & Fail & Fail & Fail & Fail \\
        \hline
        Relational & Go to the bottle on top of the table & Pass & Pass & Pass & Pass \\
        \hline
        Relational & Go to the TV closest to the person & Pass & Pass & Pass & Pass \\
        \hline
        Relational & Go to the person wearing a blue shirt & Pass & Fail & Fail & Fail \\
        \hline
        Relational & Run to to the chair with the blue coat & Pass & Fail & Fail & Fail \\
        \hline
        Relational & Move to the backpack on the chair & Fail & Fail & Fail & Fail \\
        \hline
        Relational & Move to the black backpack on the chair & Fail & Fail & Fail & Fail \\
        \hline
        Relational & Drive to the chair with the backpack on it that is not red & Fail & Fail & Fail & Fail \\
        \hline
        Contextual & Find something that can help firefighters & Pass & Pass & Pass & Pass \\
        \hline
        Contextual & Go to something that can clean the dirty floor & Fail & Fail & Fail & Fail \\
        \hline
        Contextual & I am thirsty. Walk to somewhere this can be fixed & Fail & Fail & Fail & Fail \\
        \hline
        Contextual & Go to a place where I can watch a movie & Pass & Pass & Pass & Pass \\
        \hline
        Contextual & Drive to a place where I can watch a video & Fail & Fail & Fail & Fail \\
        \hline
        
    \end{tabular}
\end{table}

\begin{table}
    \caption{Scene-03: Outdoor}
    \label{outdoor_sentences}
    \rowcolors{2}{white}{gray!30}
    \begin{tabular}{|p{1.5cm}|p{7cm}|p{0.9cm}|p{0.9cm}|p{0.9cm}|p{0.9cm}|}
        \hline
        \textbf{Category} & \textbf{Scene} & \textbf{Code} & \textbf{OD} & \textbf{WP} & \textbf{Path\& Exec} \\
        \hline \hline
        Generic & Wander to fire hydrant & Pass & Pass & Pass & Pass \\
        \hline
        Generic & Step towards the grill & Pass & Fail & Fail & Fail \\
        \hline
        Generic & Walk to the skateboard & Pass & Pass & Pass & Pass \\
        \hline
        Generic & Walk to the bike & Pass & Pass & Fail & Fail \\
        \hline
        Generic & Walk to the bike & Pass & Pass & Pass & Pass \\
        \hline
        Generic & Go to bike rack & Pass & Pass & Pass & Pass \\
        \hline
        Generic & Go to the sign & Pass & Pass & Fail & Fail \\
        \hline
        Generic & Go to the bench & Pass & Pass & Pass & Pass \\
        \hline
        Specific & Sashay to the stop sign & Pass & Pass & Pass & Pass \\
        \hline
        Specific & Go to the basketball hoop & Pass & Pass & Pass & Pass \\
        \hline
        Specific & Roam towards the blue car & Pass & Pass & Pass & Pass \\
        \hline
        Specific & Trot towards the red bag & Pass & Pass & Pass & Pass \\
        \hline
        Specific & Go to the red object generically & Fail & Fail & Fail & Fail \\
        \hline
        Relational & Proceed to the middle cone & Pass & Pass & Pass & Pass \\
        \hline
        Relational & Journey to the tree next to the backpack & Fail & Fail & Fail & Fail \\
        \hline
        Relational & Trek to the backpack by the tree & Pass & Pass & Pass & Pass \\
        \hline
        Contextual & A firefighter needs water. Walk to a source of water & Pass & Fail & Fail & Fail \\
        \hline
        Contextual & Head towards something that can help firefighters & Pass & Pass & Pass & Pass \\
        \hline
        Contextual & You are a dog that needs to mark its territory. Go to a place that can do this & Pass & Pass & Pass & Pass \\
        \hline
        Contextual & You are carrying trash, Find somewhere to dump & Pass & Pass & Pass & Pass \\
        \hline
        Contextual & Find me something to do a kick flip on & Pass & Pass & Pass & Pass \\
        \hline
        Contextual & I want to shoot some hoops. Take me there & Fail & Fail & Fail & Fail \\
        \hline
        Contextual & Move to a faster mode of transportation & Pass & Pass & Pass & Pass \\
        \hline
        Contextual & Head to the fastest mode of transportation & Pass & Pass & Pass & Pass \\
        \hline
    \end{tabular}
\end{table}

\begin{table}
    \caption{Scene-04: Courtyard}
    \label{courtyard_sentences}
    \rowcolors{2}{white}{gray!30}
    \begin{tabular}{|p{1.5cm}|p{7cm}|p{0.9cm}|p{0.9cm}|p{0.9cm}|p{0.9cm}|}
        \hline
        \textbf{Category} & \textbf{Scene} & \textbf{Code} & \textbf{OD} & \textbf{WP} & \textbf{Path\& Exec} \\
        \hline \hline
        Generic & Run to the door & Pass & Pass & Pass & Pass \\
        \hline
        Generic & Run towards the backpack & Pass & Pass & Pass & Pass \\
        \hline
        Generic & Drive to the wagon & Pass & Pass & Pass & Pass \\
        \hline
        Generic & Navigate to the stairs & Pass & Pass & Pass & Pass \\
        \hline
        Specific & Proceed towards the garbage can on the right & Pass & Pass & Pass & Pass \\
        \hline
        Specific & Stroll to the recycle bin on the left & Pass & Pass & Fail & Fail \\
        \hline
        Specific & Sprint to the picnic table & Pass & Pass & Pass & Pass \\
        \hline
        Relational & Go to the bench with water container & Pass & Pass & Pass & Pass \\
        \hline
        Relational & Walk to the bench with nothing on it & Pass & Pass & Pass & Pass \\
        \hline
        Relational & Proceed to the bench with most objects on it & Fail & Fail & Fail & Fail \\
        \hline
        Relational & Move towards the backpack farthest from the bench & Pass & Fail & Fail & Fail \\
        \hline
        Relational & Head towards the middle cone & Fail & Fail & Fail & Fail \\
        \hline
        Relational & Head towards to middle cone in the row of cones & Pass & Pass & Pass & Pass \\
        \hline
        Relational & Go to the table with only 1 chair & Fail & Fail & Fail & Fail \\
        \hline
        Relational & Go to the table with only 1 chair. There are multiple groups of chairs multiple tables & Pass & Pass & Pass & Pass \\
        \hline
        Relational & Step towards the column closest to the cart & Pass & Pass & Pass & Pass \\
        \hline
        Relational & Move to the largest group of benches & Pass & Pass & Pass & Pass \\
        \hline
        Relational & Walk towards the table with the umbrella & Fail & Fail & Fail & Fail \\
        \hline
        Relational & Walk towards the table with the umbrella & Pass & Pass & Pass & Pass \\
        \hline
        Relational & Drive to the table without any chairs & Pass & Pass & Pass & Pass \\
        \hline
        Relational & Walk to black table with 6 chairs & Fail & Fail & Fail & Fail \\
        \hline
        Relational & Walk to the table with most chairs & Pass & Pass & Pass & Pass \\
        \hline
        Relational & Navigate to the table with the backpack & Fail & Fail & Fail & Fail \\
        \hline
        Contextual & Go to nearest entrance to the building & Fail & Fail & Fail & Fail \\
        \hline
        Contextual & Go to something that you would hold open for someone elderly & Pass & Fail & Fail & Fail \\
        \hline
        Contextual & Go to something that will make it easy to carry heavy luggage & Pass & Fail & Fail & Fail \\
        \hline
        Contextual & Go to somewhere I can eat my lunch & Pass & Fail & Fail & Fail \\
        \hline
        Contextual & Go up to the second floor & Fail & Fail & Fail & Fail \\
        \hline
        Contextual & Go find something to climb & Fail & Fail & Fail & Fail \\
        \hline
        Contextual & Run upstairs & Pass & Pass & Pass & Pass \\
        \hline
        Contextual & Find me somewhere to park my bike & Pass & Pass & Pass & Pass \\
        \hline
    \end{tabular}
\end{table}

All the prompts used in classroom environment for testing the two different input representations A \& B are presented in Table \ref{classroom_sentences}.
\begin{table}
    \caption{Scene-05: Classroom}
    \label{classroom_sentences}
    \rowcolors{2}{white}{gray!30}
    \begin{tabular}{|p{1.5cm}|p{9cm}|p{1cm}|p{1cm}|}
        \hline
        \textbf{Category} & \textbf{Sentence} & \textbf{A} & \textbf{B} \\
        \hline \hline
        Generic & Go to the backpack & Pass & Pass \\
        \hline
        Generic & Move towards the backpack & Pass & Pass \\
        \hline
        Generic & Drive to the backpack & Pass & Pass \\
        \hline
        Generic & Run towards the backpack & Pass & Pass \\
        \hline
        Generic & Go to the cone & Pass & Pass \\
        \hline
        Generic & Go to conical traffic delineator & Pass & Pass \\
        \hline
        Generic & Go to the trash can & Pass & Pass \\
        \hline
        Generic & Walk to the whiteboard & Pass & Pass \\
        \hline
        Generic & Proceed to the broom & Pass & Pass \\
        \hline
        Generic & Trek towards the wagon & Pass & Pass \\
        \hline
        Generic & Find paper towels & Pass & Pass \\
        \hline
        Generic & Go to the outlet & Pass & Pass \\
        \hline
        Specific & Go to the red backpack & Pass & Pass \\
        \hline
        Specific & Go to the black backpack & Pass & Pass \\
        \hline
        Specific & Navigate to the backpack on the left & Pass & Pass \\
        \hline
        Specific & Drive to the backpack on the right & Pass & Pass \\
        \hline
        Specific & Go to the whiteboard in front of you & Pass & Pass \\
        \hline
        Specific & Move to the whiteboard on your right & Pass & Pass \\
        \hline
        Specific & Move to the whiteboard on the right & Pass & Fail \\
        \hline
        Specific & Go to the backpack on your right & Pass & Pass \\
        \hline
        Specific & Walk to the backpack on the left & Pass & Fail \\
        \hline
        Specific & Go to the leftmost backpack on the right & Pass & Fail \\
        \hline
        Specific & Go to the orange cone on your right & Fail & Pass \\
        \hline
        Specific & Go to the middle outlet & Pass & Fail \\
        \hline
        Relational & Go to backpack to the right of the red backpack & Pass & Pass \\
        \hline
        Relational & Drive to the backpack that is to the left of the black backpack & Pass & Fail \\
        \hline
        Relational & Walk to the bag that is next to the black bag & Fail & Fail \\
        \hline
        Relational & Move towards the backpack under the whiteboard & Pass & Pass \\
        \hline
        Relational & Walk to the backpack on the chair & Pass & Pass \\
        \hline
        Relational & Go to the chair with the backpack & Pass & Fail \\
        \hline
        Relational & Walk to the backpack on top of the chair & Pass & Fail \\
        \hline
        Relational & Run to the rightmost backpack & Pass & Pass \\
        \hline
        Relational & Walk to the leftmost backpack & Pass & Pass \\
        \hline
        Relational & Go to the middle chair & Pass & Fail \\
        \hline
        Relational & Go to the leftmost backpack on your right & Fail & Pass \\
        \hline
        Relational & Go to the middle chair in the row of chairs & Pass & Pass \\
        \hline
        Relational & Go to the backpack to the left of the cone & Pass & Fail \\
        \hline
        Relational & Go the cone to the left of the backpack & Pass & Pass \\
        \hline
        Relational & Go to the second chair from the left & Pass & Fail \\
        \hline
        Contextual & Go to somewhere I can sit down & Pass & Pass \\
        \hline
        Contextual & Find a place for me to rest & Pass & Fail \\
        \hline
        Contextual & Go to somewhere I can speak from & Pass & Fail \\
        \hline
        Contextual & Find a place to store cleaning supplies & Pass & Fail \\
        \hline
        Contextual & Find me something to write on & Pass & Pass \\
        \hline
        Contextual & My friend has question. Go to somewhere you can explain the answer to him & Fail & Fail \\
        \hline
        Contextual & I spilled a lot of sand. Find me something to pick up my mess & Pass & Pass \\
        \hline
        Contextual & Walk to something I can put my laptop in & Fail & Fail \\
        \hline
        Contextual & I spilled water. Find me something to clean this up & Pass & Fail \\
        \hline
        Contextual & Go to somewhere I can google something & Pass & Pass \\
        \hline
        Contextual & Go to somewhere I can charge my phone & Pass & Pass \\
        \hline
    \end{tabular}
\end{table}
\newpage
\clearpage
\subsection{Code Prompt}
Navigation Prompt for the concatenated single image input. We leverage the original prompts presented in \cite{suris2023vipergpt}.

\begin{lstlisting}[language=Python ]
import math

class ImagePatch:
    """A Python class containing a crop of an image centered around a particular object, as well as relevant information.
    Attributes
    ----------
    cropped_image : array_like
        An array-like of the cropped image taken from the original image.
    left, lower, right, upper : int
        An int describing the position of the (left/lower/right/upper) border of the crop's bounding box in the original image.
    frame: name of camera frame

    Methods
    -------
    find(object_name: str)->List[ImagePatch]
        Returns a list of new ImagePatch objects containing crops of the image centered around any objects found in the
        image matching the object_name.
    exists(object_name: str)->bool
        Returns True if the object specified by object_name is found in the image, and False otherwise.
    verify_property(property: str)->bool
        Returns True if the property is met, and False otherwise.
    best_text_match(option_list: List[str], prefix: str)->str
        Returns the string that best matches the image.
    simple_query(question: str=None)->str
        Returns the answer to a basic question asked about the image. If no question is provided, returns the answer to "What is this?".
    llm_query(question: str, long_answer: bool)->str
        References a large language model (e.g., GPT) to produce a response to the given question. Default is short-form answers, can be made long-form responses with the long_answer flag.
    compute_depth()->float
        Returns the median depth of the image crop.
    crop(left: int, lower: int, right: int, upper: int)->ImagePatch
        Returns a new ImagePatch object containing a crop of the image at the given coordinates.
    """

    def __init__(self, image, left: int = None, lower: int = None, right: int = None, upper: int = None, frame = None):
        """Initializes an ImagePatch object by cropping the image at the given coordinates and stores the coordinates as
        attributes. If no coordinates are provided, the image is left unmodified, and the coordinates are set to the
        dimensions of the image.
        Parameters
        -------
        image : array_like
            An array-like of the original image.
        left, lower, right, upper : int
            An int describing the position of the (left/lower/right/upper) border of the crop's bounding box in the original image.
        """
        if left is None and right is None and upper is None and lower is None:
            self.cropped_image = image
            self.left = 0
            self.lower = 0
            self.right = image.shape[2]  # width
            self.upper = image.shape[1]  # height
        else:
            self.cropped_image = image[:, lower:upper, left:right]
            self.left = left
            self.upper = upper
            self.right = right
            self.lower = lower

        self.width = self.cropped_image.shape[2]
        self.height = self.cropped_image.shape[1]

        self.horizontal_center = (self.left + self.right) / 2
        self.vertical_center = (self.lower + self.upper) / 2

        self.frame = frame

    def find(self, object_name: str) -> List[ImagePatch]:
        """Returns a list of ImagePatch objects matching object_name contained in the crop if any are found.
        Otherwise, returns an empty list.
        Parameters
        ----------
        object_name : str
            the name of the object to be found

        Returns
        -------
        List[ImagePatch]
            a list of ImagePatch objects matching object_name contained in the crop

        Examples
        --------
        >>> # return the foo
        >>> def execute_command(image) -> List[ImagePatch]:
        >>>     image_patch = ImagePatch(image)
        >>>     foo_patches = image_patch.find("foo")
        >>>     return foo_patches
        """
        return find_in_image(self.cropped_image, object_name)

    def exists(self, object_name: str) -> bool:
        """Returns True if the object specified by object_name is found in the image, and False otherwise.
        Parameters
        -------
        object_name : str
            A string describing the name of the object to be found in the image.

        Examples
        -------
        >>> # Are there both foos and garply bars in the photo?
        >>> def execute_command(image)->str:
        >>>     image_patch = ImagePatch(image)
        >>>     is_foo = image_patch.exists("foo")
        >>>     is_garply_bar = image_patch.exists("garply bar")
        >>>     return bool_to_yesno(is_foo and is_garply_bar)
        """
        return len(self.find(object_name)) > 0

    def verify_property(self, object_name: str, visual_property: str) -> bool:
        """Returns True if the object possesses the visual property, and False otherwise.
        Differs from 'exists' in that it presupposes the existence of the object specified by object_name, instead checking whether the object possesses the property.
        Parameters
        -------
        object_name : str
            A string describing the name of the object to be found in the image.
        visual_property : str
            A string describing the simple visual property (e.g., color, shape, material) to be checked.

        Examples
        -------
        >>> # Do the letters have blue color?
        >>> def execute_command(image) -> str:
        >>>     image_patch = ImagePatch(image)
        >>>     letters_patches = image_patch.find("letters")
        >>>     # Question assumes only one letter patch
        >>>     return bool_to_yesno(letters_patches[0].verify_property("letters", "blue"))
        """
        return verify_property(self.cropped_image, object_name, property)

    def best_text_match(self, option_list: List[str]) -> str:
        """Returns the string that best matches the image.
        Parameters
        -------
        option_list : str
            A list with the names of the different options
        prefix : str
            A string with the prefixes to append to the options

        Examples
        -------
        >>> # Is the foo gold or white?
        >>> def execute_command(image)->str:
        >>>     image_patch = ImagePatch(image)
        >>>     foo_patches = image_patch.find("foo")
        >>>     # Question assumes one foo patch
        >>>     return foo_patches[0].best_text_match(["gold", "white"])
        """
        return best_text_match(self.cropped_image, option_list)

    def simple_query(self, question: str = None) -> str:
        """Returns the answer to a basic question asked about the image. If no question is provided, returns the answer
        to "What is this?". The questions are about basic perception, and are not meant to be used for complex reasoning
        or external knowledge.
        Parameters
        -------
        question : str
            A string describing the question to be asked.

        Examples
        -------

        >>> # Which kind of baz is not fredding?
        >>> def execute_command(image) -> str:
        >>>     image_patch = ImagePatch(image)
        >>>     baz_patches = image_patch.find("baz")
        >>>     for baz_patch in baz_patches:
        >>>         if not baz_patch.verify_property("baz", "fredding"):
        >>>             return baz_patch.simple_query("What is this baz?")

        >>> # What color is the foo?
        >>> def execute_command(image) -> str:
        >>>     image_patch = ImagePatch(image)
        >>>     foo_patches = image_patch.find("foo")
        >>>     foo_patch = foo_patches[0]
        >>>     return foo_patch.simple_query("What is the color?")

        >>> # Is the second bar from the left quuxy?
        >>> def execute_command(image) -> str:
        >>>     image_patch = ImagePatch(image)
        >>>     bar_patches = image_patch.find("bar")
        >>>     bar_patches.sort(key=lambda x: x.horizontal_center)
        >>>     bar_patch = bar_patches[1]
        >>>     return bar_patch.simple_query("Is the bar quuxy?")
        """
        return simple_query(self.cropped_image, question)

    def compute_depth(self):
        """Returns the median depth of the image crop
        Parameters
        ----------
        Returns
        -------
        float
            the median depth of the image crop

        Examples
        --------
        >>> # the bar furthest away
        >>> def execute_command(image)->ImagePatch:
        >>>     image_patch = ImagePatch(image)
        >>>     bar_patches = image_patch.find("bar")
        >>>     bar_patches.sort(key=lambda bar: bar.compute_depth())
        >>>     return bar_patches[-1]
        """
        depth_map = compute_depth(self.cropped_image)
        return depth_map.median()

    def crop(self, left: int, lower: int, right: int, upper: int) -> ImagePatch:
        """Returns a new ImagePatch cropped from the current ImagePatch.
        Parameters
        -------
        left, lower, right, upper : int
            The (left/lower/right/upper)most pixel of the cropped image.
        -------
        """
        return ImagePatch(self.cropped_image, left, lower, right, upper, self.frame)

    def overlaps_with(self, left, lower, right, upper):
        """Returns True if a crop with the given coordinates overlaps with this one,
        else False.
        Parameters
        ----------
        left, lower, right, upper : int
            the (left/lower/right/upper) border of the crop to be checked

        Returns
        -------
        bool
            True if a crop with the given coordinates overlaps with this one, else False

        Examples
        --------
        >>> # black foo on top of the qux
        >>> def execute_command(image) -> ImagePatch:
        >>>     image_patch = ImagePatch(image)
        >>>     qux_patches = image_patch.find("qux")
        >>>     qux_patch = qux_patches[0]
        >>>     foo_patches = image_patch.find("black foo")
        >>>     for foo in foo_patches:
        >>>         if foo.vertical_center > qux_patch.vertical_center
        >>>             return foo
        """
        return self.left <= right and self.right >= left and self.lower <= upper and self.upper >= lower

def best_image_match(list_patches: List[ImagePatch], content: List[str], return_index=False) -> Union[ImagePatch, int]:
    """Returns the patch most likely to contain the content.
    Parameters
    ----------
    list_patches : List[ImagePatch]
    content : List[str]
        the object of interest
    return_index : bool
        if True, returns the index of the patch most likely to contain the object

    Returns
    -------
    int
        Patch most likely to contain the object
    """
    return best_image_match(list_patches, content, return_index)


def distance(patch_a: ImagePatch, patch_b: ImagePatch) -> float:
    """
    Returns the distance between the edges of two ImagePatches. If the patches overlap, it returns a negative distance
    corresponding to the negative intersection over union.

    Parameters
    ----------
    patch_a : ImagePatch
    patch_b : ImagePatch

    Examples
    --------
    # Return the qux that is closest to the foo
    >>> def execute_command(image):
    >>>     image_patch = ImagePatch(image)
    >>>     qux_patches = image_patch.find('qux')
    >>>     foo_patches = image_patch.find('foo')
    >>>     foo_patch = foo_patches[0]
    >>>     qux_patches.sort(key=lambda x: distance(x, foo_patch))
    >>>     return qux_patches[0]
    """
    return distance(patch_a, patch_b)


def bool_to_yesno(bool_answer: bool) -> str:
    return "yes" if bool_answer else "no"


def coerce_to_numeric(string):
    """
    This function takes a string as input and returns a float after removing any non-numeric characters.
    If the input string contains a range (e.g. "10-15"), it returns the first value in the range.
    """
    return coerce_to_numeric(string)

### Nav Client
"""
Navigates an agent to an object in an imgae given its center coordinates
Parameters
    ----------
    x : x coordinate of the center of the object
    y : y coordinate of the center of the object
"""
 Examples
        -------
        >>> # Go to the blue foo.
        >>> def execute_command(image) 
        >>>     image_patch = ImagePatch(image)
        >>>     foo_patches = image_patch.find("foo")
        >>>     # Verify visual property
        >>>     blue_color_patches = []
        >>>     for foo_patch in foo_patches:
        >>>         if verify_property(blue, "color")
        >>>            blue_color_patches.append(foo_patch)
        >>>     inputs = (blue_color_patches[0].horizontal_center, blue_color_patches[0].vertical_center)
        >>>     return {'function': 'nav_function', 'inputs': inputs, 'box': [blue_color_patches[0].left, blue_color_patches[0].lower, blue_color_patches[0].right, blue_color_patches.upper]}                                                   
        """
def navigate_to_object(double x, double y):

"""



Write a function using Python and/or the ImagePatch class (above) that could be executed to provide an answer to the query by calling one of the functions in the Nav Client (navigate_to_object). 
Note using the ImagePatch is not required for all queries. If the query is not relevant to navigation, return None for the function and a string describing the problem. 

Consider the following guidelines:
- Use base Python (comparison, sorting) for basic logical operations, left/right/up/down, math, etc.
- Use the llm_query function to access external information and answer informational questions not concerning the image.
- All properties such as color should be verified using the verify_property function. "go to the blue foo" implies "go to foo if foo is blue"
- The output of this function should be a dict {function: 'function in the nav client', inputs: 'inputs to the chosen nav client function', box: [left, lower, right, upper]}, or for an error {function: 'None', error: 'message describing the problem'}
- If more than one object is found pick the best match

Query: INSERT_QUERY_HERE

\end{lstlisting}

\end{document}